\documentclass{article}

\usepackage{microtype}
\usepackage{graphicx}
\usepackage{subfigure}
\usepackage{booktabs}
\usepackage{multirow}

\usepackage[accepted,nohyperref]{icml2019}
\usepackage{amsmath}
\usepackage{amsfonts,amssymb}
\usepackage{bm}
\usepackage{graphicx}

\icmltitlerunning{Convolutional Poisson Gamma Belief Network}

\begin{document}

\twocolumn[
\icmltitle{Convolutional Poisson Gamma Belief Network}

\begin{icmlauthorlist}
\icmlauthor{Chaojie Wang}{xd}
\icmlauthor{Bo Chen}{xd}
\icmlauthor{Sucheng Xiao}{xd}
\icmlauthor{Mingyuan Zhou}{ut}
\end{icmlauthorlist}

\icmlaffiliation{xd}{National Laboratory of Radar Signal Processing, Collaborative Innovation Center of Information Sensing and Understanding, Xidian University, Xi'an, Shaanxi, China.}
\icmlaffiliation{ut}{McCombs School of Business, The University of Texas at Austin, Austin, Texas 78712, USA}
\icmlcorrespondingauthor{Bo Chen}{bchen@mail.xidian.edu.cn}
\icmlkeywords{Bayesian Deep Learning, Generative Models}
\vskip 0.3in
]
\printAffiliationsAndNotice{} %

\begin{abstract}
For text analysis, one often resorts to a lossy representation that either completely ignores word order or embeds each word as a low-dimensional dense feature vector. In this paper, we propose convolutional Poisson factor analysis (CPFA) that directly operates on a lossless representation that processes the words in each document as a sequence of high-dimensional one-hot vectors. To boost its performance, we further propose the convolutional Poisson gamma belief network (CPGBN) that couples CPFA with the gamma belief network via a novel probabilistic pooling layer. CPFA forms words into phrases and captures very specific phrase-level topics, and CPGBN further builds a hierarchy of increasingly more general phrase-level topics. For efficient inference, we develop both a Gibbs sampler and a Weibull distribution based convolutional variational auto-encoder. Experimental results demonstrate that CPGBN can extract high-quality text latent representations that capture the word order information, and hence can be leveraged as a building block to enrich a wide variety of existing latent variable models that ignore word order.
\end{abstract}

\section{Introduction}
\label{Introduction}
A central task in text analysis and language modeling is to effectively represent the documents to capture their underlying semantic structures.
A basic idea is to represent the words appearing in a document with a sequence of one-hot vectors, where the vector dimension is the size of the vocabulary.
This preserves all textual information but results in a collection of extremely large and sparse matrices for a text corpus. %
Given the memory and computation constraints, it is very challenging to directly model this lossless representation. Thus existing methods often resort to simplified lossy representations that either completely ignore word order \cite{blei2003latent}, or embed the words into a lower dimensional feature space \cite{mikolov2013distributed}.

Ignoring word order, each document is simplified as a bag-of-words count vector, the $v$th element of which represents how many times the $v$th vocabulary term appears in that document.
With a text corpus simplified as a term-document frequency count matrix, a wide array of latent variable models (LVMs) have been proposed for text analysis \cite{LSA,LSI,NMF,blei2003latent,hinton2009replicated,BNBP_PFA_AISTATS2012}.
Extending ``shallow'' probabilistic topic models such as latent Dirichlet allocation (LDA) \cite{blei2003latent} and Poisson factor analysis (PFA) \cite{BNBP_PFA_AISTATS2012}, steady progress has been made in inferring multi-stochastic-layer deep latent representations for text analysis \cite{gan2015scalable,zhou2016augmentable,ranganath2015deep,zhang2018whai}.
Despite the progress, completely ignoring word order could still be particularly problematic on some common text-analysis tasks, such as spam detection and sentiment analysis \cite{pang2002thumbs,tang2014learning}.

To preserve word order, a common practice is to first convert each word in the vocabulary from a high-dimensional sparse one-hot vector into a low-dimensional dense word-embedding vector.
The word-embedding vectors can be either trained as part of the learning \cite{kim2014, kalchbrenner2014a}, or pre-trained by some other methods on an additional large corpus \cite{mikolov2013distributed}.
Sequentially ordered word embedding vectors have been successfully combined with deep neural networks to address various problems in text analysis and language modeling.
A typical combination method is to use the word-embedding layer as part of a recurrent neural network (RNN), especially long short-term memory (LSTM) and its variants \cite{hochreiter1997long,chung2014empirical}, achieving great success in numerous tasks that heavily rely on having high-quality sentence representation.
Another popular combination method is to apply a convolutional neural network (CNN) \cite{lecun1998gradient-based} directly to the embedding representation, treating the word embedding layer as an image input; it has been widely used in systems for entity search, sentence modeling, product feature mining, and so on \cite{xu2013convolutional,weston2014}.

In this paper, we first propose convolutional PFA (CPFA) that directly models the documents, each of which is represented without information loss as a sequence of one-hot vectors.
We then boot its performance by coupling it with the gamma belief network (GBN) of \citet{zhou2016augmentable}, a multi-stochastic-hidden layer deep generative model, via a novel probabilistic document-level pooling layer. We refer to the CPFA and GBN coupled model
as convolutional Poisson GBN (CPGBN). %
To the best of our knowledge, CPGBN is the first unsupervised probabilistic convolutional model that infers multi-stochastic-layer latent variables for documents represented without information loss.
Its hidden layers can be jointly trained with an upward-downward Gibbs sampler; this makes its inference different from greedy layer-wise training \cite{lee2009unsupervised, chen2013deep}. In each Gibbs sampling iteration, the main computation is embarrassingly parallel and hence will be accelerated with Graphical Process Units (GPUs).
We also develop a Weibull distribution based convolutional variational auto-encoder to provide amortized variational inference, which further accelerates both training and testing for large corpora.
Exploiting the multi-layer structure of CPGBN, we further propose a supervised CPGBN (sCPGBN), which combines the representation power of CPGBN for topic modeling and the discriminative power of deep neural networks (NNs) under a principled probabilistic framework.
We show that the proposed models achieve state-of-art results in a variety of text-analysis tasks.

\section{Convolutional Models for Text Analysis} %
Below we introduce CPFA and then develop a probabilistic document-level pooling method to couple CPFA with GBN, which further serves as the decoder of a Weibull distribution based convolutional variational auto-encoder (VAE).

\subsection{Convolutional Poisson Factor Analysis}
Denote $V$ as the vocabulary and let {${D_j} =( {x_{j1}},...,{x_{jL_j}})$} represent the $L_j$ sequentially ordered words of the $j$th document, which can be represented as a sequence of one-hot vectors. %
For example, with vocabulary $V = \{$\emph{``don't'',``hate'',``I'',``it'',``like''}$\}$, document ${D_j}=$ (\emph{``I'',``like'',``it''}) can be represented as $\bm{X}_j = [\bm{x}_{j1},\bm{x}_{j2},\bm{x}_{j3}] \in \{0,1\}^{\left| V \right| \times L_j}$, where $\bm{x}_{j1} = (0,0,1,0,0)'$, $\bm{x}_{j2} = (0,0,0,0,1)'$, and $\bm{x}_{j3} = (0,0,0,1,0)'$ are one-hot column vectors.
Let us denote $x_{jvl}=\bm{X}_j(v,l)$, which is one if and only if word $l$ of document $j$ matches term $v$ of the vocabulary.

To exploit a rich set of tools developed for count data analysis \citep{BNBP_PFA_AISTATS2012,zhou2016augmentable}, we first link these sequential binary vectors to sequential count vectors via the Bernoulli-Poisson link \citep{EPM_AISTATS2015}.
More specifically, we link each $x_{jvl}$ to a latent count as $x_{jvl}=\mathbf{1}(m_{jvk}>0)$, where $m_{jvl}\in\mathbb{Z}:=\{0,1,\ldots\}$, and factorize the matrix $\bm M_j=\{m_{jvl}\}_{v,l}\in \mathbb{Z}^{\left| V \right| \times L_j}$ under the Poisson likelihood.
Distinct from vanilla PFA \citep{BNBP_PFA_AISTATS2012} where the columns of the matrix are treated as conditionally independent, here we introduce convolution into the hierarchical model to capture the sequential dependence between the columns. We construct the hierarchical model of CPFA as
\begin{equation}
 \,\,\begin{matrix} %
 \bm{X}_j = \mathbf{1}(\bm{M}_j >0),~ {{\bm{M}}_{j}}\sim \mbox{Pois}(\sum\nolimits_{{k=1}}^{{K}} {{\bm{D}}_{{k}}*{\bm{w}}_{jk}} ),\vspace{1.1mm}\\ %
 {\bm{w}}_{jk} \sim \mbox{Gam}(r_{{k}},1/{c_j}),~~
 \bm{D}_{k}(:) \sim \mbox{Dir}(\eta \bm 1_{|V| F}),
 \end{matrix}
\label{eq_CPFA}
\end{equation}
where $*$ denotes a convolution operator, ${\mathbb{R}_{+}}:=\{x:x\ge 0\}$, $\bm{D}_k =(\bm d_{k1},\ldots,\bm d_{kF})\in \mathbb{R}_+^{|V|\times F}$ is the $k$th convolutional filter/factor/topic whose filter width is $F$, $\bm d_{kf}=(d_{k1f},\ldots,d_{k|V|f})'$, and
$\bm{D}_{k}(:)=(\bm d_{k1}',\ldots,\bm d_{kF}')'\in \mathbb{R}_+^{|V|F}$; %
the latent count matrix ${\bm{M}_j}$ is factorized into the summation of $K$ equal-sized latent count matrices, the Poisson rates of the $k$th of which are obtained by convolving $\bm{D}_k$ with its corresponding gamma distributed feature representation ${{\bm{w}}_{jk}} \in \mathbb{R}_ + ^{S_j}$, where $S_j:=L_j-F+1$.
To complete the hierarchical model, we let $r_k\sim \mbox{Gamma}(1/{K},1/c_0)$ and ${c_j}\sim \mbox{Gamma}({e_0},1/{f_0})$.
Note as in \citet{zhou2016augmentable}, we may consider $K$ as the truncation level of a gamma process, which allows the number of needed factors to be inferred from the data as long as $K$ is set sufficiently large.

We can interpret $d_{kvf}:=\bm{D}_k(v,f)$ as the probability that the $v$th term in the vocabulary appears at the $f$th temporal location for the $k$th latent topic, and expect each $\bm{D}_k$ to extract both global cooccurrence patterns, such as common topics, and local temporal structures, such as common $n$-gram phrases, where $n\le F$, from the text corpus.
Note the convolution layers of CPFA convert text regions of size $F$ ($e.g.$,``am so happy'' with $F =3$) to feature vectors, directly learning the embedding of text regions without going through a separate learning for word embedding.
Thus CPFA provides a potential solution for distinguishing polysemous words according to their neighboring words.
The length of the representation weight vector {${{\bm{w}}_{jk}}$} in our model is $S_j=L_j - {F} + 1$, which varies with the document length~$L_j$.
This differs CPFA from traditional convolutional models with a fixed feature map size (\citealp{zhangdeconvolutional,miao2018direct,min2019a}), which requires either heuristic cropping or zero-padding.

\subsection{Convolutional Poisson Gamma Belief Network}
There has been significant recent interest in inferring multi-stochastic-layer deep latent representations for text analysis in an unsupervised manner \cite{gan2015scalable,
zhou2016augmentable,ranganath2015deep,wang2018multimodal, zhang2018whai}, where word order is ignored.
The key intuition behind these models, such as GBN \citep{zhou2016augmentable}, is that words frequently co-occurred in the same document can form specific word-level topics in shallow layers; as the depth of the network increases, frequently co-occurred topics can form more general ones.
Here, we propose a model to preserve word order, without losing the nice hierarchical topical interpretation provided by a deep topic model.
The intuition is that by preserving word order, words can first form short phrases; frequently co-occurred short phrases can then be combined to form specific phrase-level topics; and these specific phrase-level topics can form increasingly more general phrase-level topics when moving towards deeper layers.

\begin{figure}
 \centering
 \includegraphics[width=7.8cm]{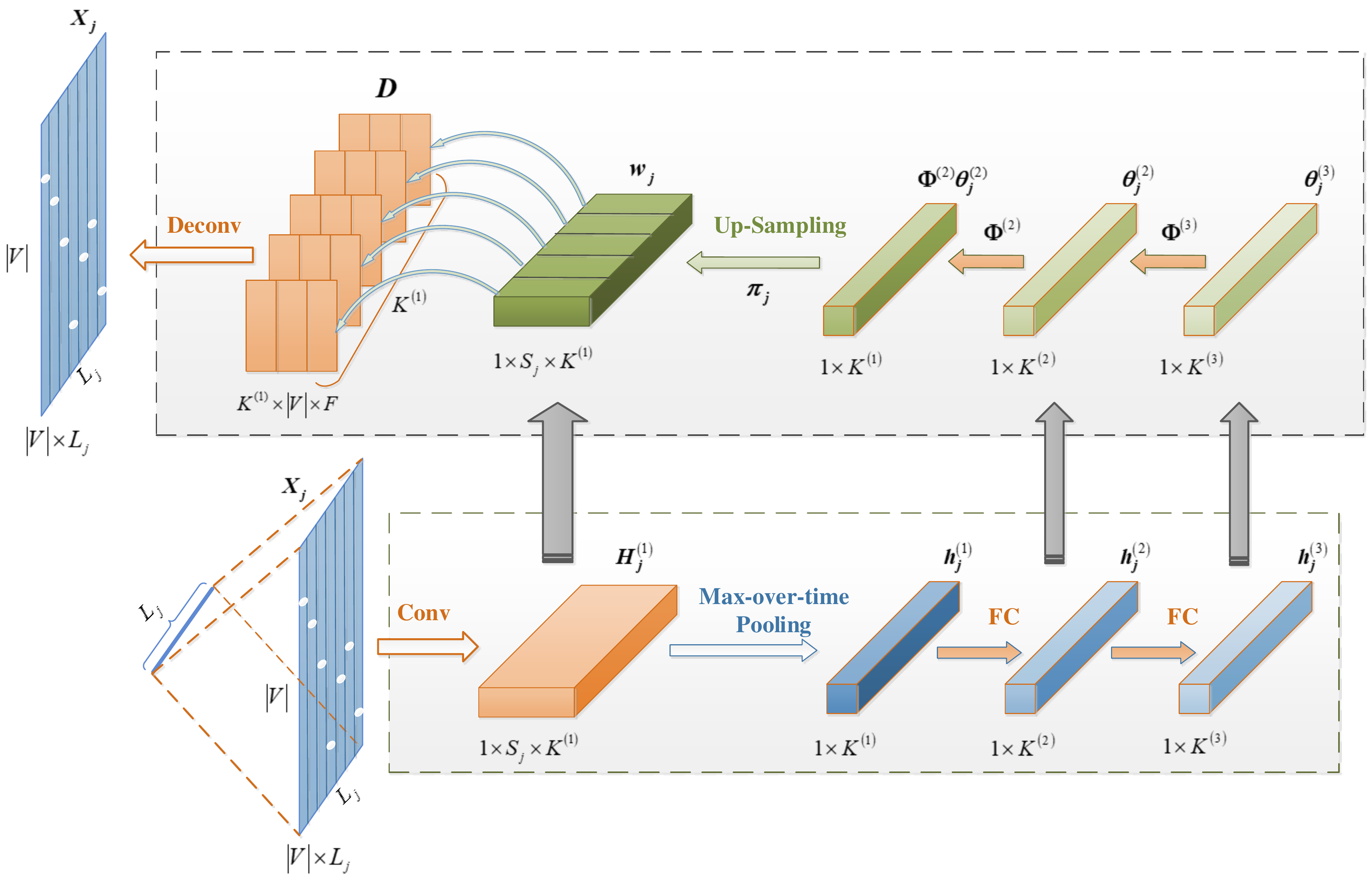} %
 \caption{The proposed CPGBN (upper part) and its corresponding convolutional variational inference network (lower part).
 }
 \label{structure} %
\end{figure}

As in Fig. \ref{structure}, we couple CPFA in \eqref{eq_CPFA} with GBN to construct CPGBN, whose generative model with $T$ hidden layers, from top to bottom, is expressed as
\begin{equation}
 \small
 \begin{array}{l}
 {\bm{\theta}} _j^{(T)}\sim \mbox{Gam}(\bm{r},1/c_j^{(T + 1)}),\\
 ...,\\
 {\bm{\theta}} _j^{(t)}\sim \mbox{Gam}({{\bm{\Phi}} ^{(t + 1)}}{\bm{\theta}} _j^{(t + 1)},1/c_j^{(t + 1)}),\\
 {...,}\\
 {\bm{\theta}} _j^{(1)}\sim\mbox{Gam}({{\bm{\Phi}} ^{(2)}}{\bm{\theta}} _j^{(2)},1/c_j^{(2)}),\vspace{1.1mm}\\
 {\bm{w}}_{j{k}} = {\bm{\pi}_{jk}} \theta _{j{k}}^{(1)},~{\bm{\pi}_{jk}} \sim \mbox{Dir}\big( {\bm{\Phi}} _{{k:}}^{({\rm{2}})}{\bm{\theta}} _j^{({\rm{2}})} /S_j \bm{1}_{S_j}\big),\vspace{1.1mm}\\
 {\bm{M}}_j\sim \mbox{Pois}\big(\sum\nolimits_{{k} = 1}^{{K^{(1)}}} {{\bm{D}}_{{k}}*{\bm{w}}_{j{k}}} \big),
 \end{array}
 \normalsize
 \label{eq_PGCN}
\end{equation}
where ${\bm{\Phi}} _{{k:}}$ is the $k$th row of $\bm{\Phi}$ and superscripts indicate layers.
Note CPGBN first factorizes the latent count matrix ${{\bm{M}}_{j}^{}} \in {\mathbb{Z}^{\left| V \right| \times L_j}}$ under the Poisson likelihood into the summation of $K^{(1)}$ convolutions, the $k$th of which is between
${{\bm{D}}_{{k}}} \in \mathbb{R}_ + ^{{\left| V \right|} \times {F}}$ and weight vector ${{\bm{w}}_{j{k}}} \in \mathbb{R}_ + ^{S_j}$.
Using the relationship between the gamma and Dirichlet distributions ($e.g.$, Lemma IV.3 of \citet{zhou2012negative}), %
${{\bm{w}}_{j{k}}}=(w_{j{k}1},\ldots,w_{j{k}S_j})'=(\theta_{jk}^{(1)}\pi_{jk1},\ldots,\theta_{jk}^{(1)}\pi_{j{k}S_j})'\in\mathbb{R}^{S_j}$ in \eqref{eq_PGCN} can be equivalently generated as
\begin{equation}
 \begin{array}{l}
 w_{j{k}s}\sim \mbox{Gam}\big( {\bm{\Phi}} _{{k:}}^{(2)}{\bm{\theta}} _j^{(2)}/S_j,1/c_j^{(2)}\big),~s=1,\ldots,S_j,
 \end{array}\!\!\!
 \label{eq_pool}
\end{equation}
which could be seen as a specific probabilistic document-level pooling algorithm on the gamma shape parameter.
For $t \in \{1,...,T-1\}$, the shape parameters of the gamma distributed hidden units ${\bm{\theta}} _j^{(t)} \in \mathbb{R}_ + ^{{K^{(t)}}}$ are factorized into the product of the connection weight matrix ${{\bm{\Phi}} ^{(t+1)}} \in \mathbb{R}_ + ^{{K^{(t)}} \times {K^{(t+1)}}}$ and hidden units ${\bm{\theta}} _j^{(t+1)} \in \mathbb{R}_ + ^{{K^{(t+1)}}}$ of layer $t+1$; the top layer's hidden units ${\bm{\theta}} _j^{(T)}$ share the same $\bm{r} \in \mathbb{R}_ + ^{{K^{(T)}}}$ as their gamma shape parameters; and ${c^{(t+1)}_j}$ are gamma scale parameters. %
For scale identifiability and ease of inference, {the columns of ${\bm{D}}_{k}$} and %
${{\bm{\Phi}} ^{(t+1)}} \in \mathbb{R}_ + ^{{K^{(t)}} \times {K^{(t+1)}}}$ are restricted to have unit $L_1$ norm. To complete the hierarchical model, we let ${\bm{D}}_{{k}}(:)\sim \mbox{Dir}({\eta ^{(1)}}{{\bm{1}}_{\left| V \right|F}})$, ${\bm{\phi}} _{{k}}^{(t)}\sim \mbox{Dir}({\eta ^{(t)}}{{\bm{1}}_{{K^{(t)}}}})$, %
$r_{k} \sim \mbox{Gam}({1}/{K^{(T)}},{1})$, and $c_j^{(t+1)}\sim \mbox{Gam}({e_0},1/{f_0})$.

Examining (\ref{eq_pool}) shows CPGBN provides a probabilistic document-level pooling layer, which %
 summarizes the content coefficients ${\bm{w}}_{j{k}}$ across all word positions into $\theta_{jk}^{(1)} = \sum_{s=1}^{S_j} w_{jks}$; the hierarchical structure after $\theta_{jk}^{(1)}$ can be flexibly modified according to the deep models (not restricted to GBN) to be combined with. %
The proposed pooling layer can be trained jointly with all the other layers, making it distinct from a usual one that often cuts off the message passing from deeper layers \cite{lee2009unsupervised, chen2013deep}.
We note using pooling on the first hidden layer is related to shallow text CNNs that use document-level pooling directly after a single convolutional layer \cite{kim2014,johnson2015effective}, which often contributes to improved efficiency \cite{boureau2010a, wang2010locality-constrained}.

\subsection{Convolutional Inference Network for CPGBN}
\label{sec_ae}
To make our model both scalable to big corpora in training and fast in out-of-sample prediction, below we
 introduce a convolutional inference network, which will be used in hybrid MCMC/variational inference described in Section~\ref{Hybrid}.
Note the usual strategy of autoencoding variational inference is to construct an inference network to map the observations directly to their latent representations, and optimize the encoder and decoder by minimizing the negative evidence lower bound (ELBO) as $L_g=\sum_{j=1}^J L_g(\bm X_j) $, where
\begin{equation}
\begin{split}
 \textstyle {L_g(\bm X_j}) = \sum_{t = 2}^T {\mathbb{E}_{Q}\big[\ln \frac{{q({\bm{\theta}} _j^{(t)}\,|\,-)}}{{p({\bm{\theta}} _j^{(t)}\,|\,{{\bm{\Phi}} ^{(t + 1)}},{\bm{\theta}} _j^{(t + 1)})}}\big]}
 \\
 \textstyle + \sum_{k = 1}^{K^{(1)}} \sum_{s=1}^{S_j} \mathbb{E}_{Q}\big[\ln \frac{{q({{w}}_{jks}\,|\,-)}}{{p({{{w}} _{jks}\,|\,{{\bm{\Phi}} ^{(2)}},{\bm{\theta}}_j^{(2)})}}}\big] \\
 \textstyle - {\mathbb{E}_{Q}[\ln p({{\bm{X}}_j}\,|\,\{{\bm{D}_k},{\bm{w}}_{jk}\}_{1,K^{(1)}})]};
\end{split}\label{eq:L}
\end{equation}
following \citet{zhang2018whai}, we use the Weibull distribution to approximate the gamma distributed conditional posterior of ${{\bm{\theta}}_j^{(t)}}$, as it is reparameterizable, resembles the gamma distribution, and the Kullback--Leibler (KL) divergence from the gamma to Weibull distributions is analytic; as in %
Fig.~\ref{structure}, we construct the autoencoding variational distribution as $Q=q({\bm{w}}_{j{k}}\,|\,-)\prod_{t=2}^T q({\bm{\theta}} _j^{(t)}\,|\,-)$, where%
\begin{equation}
 \begin{array}{l}
 q({\bm{w}}_{j{k}}\,|\,-)= \mbox{Weibull}({\bm{\Sigma}} _{j{k}}^{(1)} + {\bm{\Phi}} _{{k:}}^{(2)}{\bm{\theta}} _j^{(2)},{\bm{\Lambda}} _{j{k}}^{(1)}),\vspace{1.1mm}\\
 q({\bm{\theta}} _j^{(t)}\,|\,-)=\mbox{Weibull}({\bm{\sigma}}_j^{(t)} + {{\bm{\Phi}} ^{(t + 1)}}{\bm{\theta}} _j^{(t + 1)},{\bm{\lambda}} _j^{(t)}).\!\!
 \label{eq_w_post}
 \end{array}\!\!\!\!
\end{equation}
The parameters ${\bm{\Sigma}} _j^{(1)}, ~{\bm{\Lambda}} _j^{(1)} \in {\mathbb{R}^{{K^{(1)}} \times S_j}}$ of ${\bm{w}}_j=(\bm w_{j1},\ldots,\bm w_{jK^{(1)}})' \in {\mathbb{R}^{{K^{(1)}} \times S_j}}$ are deterministically transformed from the observation ${\bm{X}}_j$ using CNNs specified as
\begin{equation}
 \begin{array}{l}
 {\bm{H}}_j^{(1)} = \mbox{relu}({\bm{C}}_1^{(1)}*{{\bm{X}}_j} + {\bm{b}}_1^{(1)}),\vspace{1.1mm}\\
 {\bm{\Sigma}}_j^{(1)} = \mbox{exp} ({\bm{C}}_2^{(1)}*\mbox{pad}({\bm{H}}_j^{(1)}) + {\bm{b}}_2^{(1)}),\vspace{1.1mm}\\
 {\bm{\Lambda}} _j^{(1)} = \mbox{exp} ({\bm{C}}_3^{(1)}*\mbox{pad}({\bm{H}}_j^{(1)}) + {\bm{b}}_3^{(1)}),
 \end{array}\notag
\end{equation}
where $\bm{b}_1^{(1)}, \bm{b}_2^{(1)}, \bm{b}_3^{(1)} \in {{\mathbb{R}}^{{K^{(1)}}}}$, ${\bm{C}}_1^{(1)} \in {{\mathbb{R}}^{{K^{(1)}} \times \left| V \right| \times F}} $, ${\bm{C}}_2^{(1)},{\bm{C}}_3^{(1)} \in {{\mathbb{R}}^{{K^{(1)}} \times {K^{(1)}}\times F}}$, ${\bm{H}} _j^{(1)} \in {\mathbb{R}^{{K^{(1)}} \times S_j}}$, and $\mbox{pad}({\bm{H}}_j^{(1)}) \in {\mathbb{R}^{{K^{(1)}} \times L_j}} $ is obtained with zero-padding;
the parameters ${\bm{\sigma}}_j^{(t)}$ and ${\bm{\lambda}} _j^{(t)}$ are transformed from ${\bm{h}}_j^{(1)} = \mbox{pool}({\bm{H}}_j^{(1)})$ specified as
\begin{equation}
 \begin{array}{l}
 {\bm{h}}_j^{(t)} = \mbox{relu}({\bm{U}}_1^{(t)}{\bm{h}}_j^{(t - 1)} + {\bm{b}}_1^{(t)}),\vspace{1.1mm}\\
 {\bm{\sigma}}_j^{(t)} = \mbox{exp} ({\bm{U}}_2^{(t)}{\bm{h}}_j^{(t)} + {\bm{b}}_2^{(t)}),\vspace{1.1mm}\\
 {\bm{\lambda}} _j^{(t)} = \mbox{exp} ({\bm{U}}_3^{(t)}{\bm{h}}_j^{(t)} + {\bm{b}}_3^{(t)}),
 \end{array} \notag
\end{equation}
where ${\bm{b}}_1^{(t)},{\bm{b}}_2^{(t)},{\bm{b}}_3^{(t)},{\bm{h}}_j^{(t)} \in {{\mathbb{R}}^{{K^{(t)}}}}$,
${\bm{U}}_1^{(t)} \in {{\mathbb{R}}^{{K^{(t)}} \times {K^{(t - 1)}}}}$, and ${\bm{U}}_2^{(t)},{\bm{U}}_3^{(t)} \in {{\mathbb{R}}^{{K^{(t)}} \times {K^{(t)}}}}$ for $t \in \{2,...,T\}$.

Further we develop sCPGBN, a supervised generalization of CPGBN, for text categorization tasks: by adding a softmax classifier on the concatenation of ${\{ {\bm{{\theta}} _j^{(t)} \}} _{1,T}}$, the loss function of the entire framework is modified as
\begin{equation}
 L = {L_g} + \xi {L_c}, \notag
\end{equation}
where ${L_c}$ denotes the cross-entropy loss and $\xi$ is used to balance generation and discrimination (\citealp{higgins2017beta-vae}).

\section{Inference}
Below we describe the key inference equations for CPFA shown in (\ref{eq_CPFA}), a single hidden-layer version of CPGBN shown in \eqref{eq_PGCN}, and provide more details in the Appendix. How the inference of CPFA, including Gibbs sampling and hybrid MCMC/autoencoding variational inference, is generalized to that of CPGBN is similar to how the inference of PFA is generalized to that of PGBN, as described in detail in \citet{zhou2016augmentable} and \citet{zhang2018whai} and omitted here for brevity.

\subsection{Gibbs Sampling}
\label{Sec_Gibbs}
Directly dealing with the whole matrix by expanding the convolution operation with Toeplitz conversion (\citealp{bojanczyk1995on}) provides a straightforward solution for the inference of convolutional models, which transforms each observation matrix ${{\bm{M}}_{j}}$ into a vector, on which the inference methods for sparse factor analysis (\citealp{carvalho2008high-dimensional,james2010sparse}) could then be applied. %
However, considering the sparsity of the document matrix consisting of one-hot vectors, directly processing these matrices without considering sparsity will bring unnecessary burden in computation and storage. Instead, we apply data augmentation under the Poisson likelihood \cite{BNBP_PFA_AISTATS2012,zhou2016augmentable} to upward propagate latent count matrices ${{\bm{M}}_{j}}$ as
\begin{equation}
 ({\bm{M}}_{j1}^{},...,{\bm{M}}_{jK}^{}\,|\, -)\sim \mbox{Multi}({{\bm{M}}_j};{{\bm{\zeta}} _{j1}},...,{{\bm{\zeta}} _{jK}}),\notag
\end{equation}
where ${{\bm{\zeta}} _{jk}} = ({{\bm{D}}_k}*{{\bm{w}}_{jk}})/(\sum\nolimits_{k = 1}^K {{{\bm{D}}_k}*{{\bm{w}}_{jk}}} )$.
Note we only need to focus on nonzero elements of ${\bm{M}}_{jk}^{} \in {{\mathbb{Z}}^{\left| V \right| \times L_j}}$.
We rewrite the likelihood function by expanding the convolution operation along the dimension of ${\bm{w}}_{jk}$ as
\begin{equation}
 \begin{array}{c}
 {m_{jkvl}}\sim \mbox{Pois}(\sum\nolimits_{{s} = 1}^{{S_j}} {{w_{jk{s}}}{d_{kv{(l-s+1)}}}} ),
 \end{array}\notag
\end{equation}
where $d_{kv(l-s+1)}:=0$ if $l-s+1\notin \{1,2,\ldots,F\}$.
Thus each nonzero element $m_{jkvl}^{{}}$ could be augmented as
\begin{equation}
 ({\bm{m}}_{jkvl}^{{}}\,|{\,m_{jkvl}^{{}}})\sim \mbox{Multi}(m_{jkvl}^{{}};{\delta _{jkvl1}},...,{\delta _{{jkvlS_j}}}),
 \label{eq_w_aug}
\end{equation}
where ${\delta _{{jkvls}}} = {w_{jk{s}}}{d_{kv{(l-s+1)}}}/\sum\nolimits_{{s} = 1}^{{S_j}} {{w_{jk{s}}}{d_{kv{(l-s+1)}}}} $ and ${\bm{m}}_{jkvl}^{{}} \in {{\mathbb{Z}}^{{S_j}}}$.
We can now decouple ${{{\bm{D}}_{{k}}}*{\bm{w}}_{jk}^{{}}}$ in (\ref{eq_CPFA}) by marginalizing out ${\bm{D}}_{{k}}$, leading to
\begin{equation}
 \textstyle \bm m_{j k \bm \cdot \bm \cdot} \sim \mbox{Pois}({{\bm{w}}_{jk}}). \notag
\end{equation}
where the symbol ``$\bm \cdot$'' denotes summing over the corresponding index and hence $\bm m_{j k \bm \cdot \bm \cdot}=\sum_{v = 1}^{\left| V \right|} \sum_{l = 1}^{{L_j}} {{{\bm{m}}_{jkvl}}}$.
Using the gamma-Poisson conjugacy, we have
\begin{equation}\textstyle
 ({{\bm{w}}_{jk}}\,|\,{-})\sim \mbox{Gam}(m_{j k \bm \cdot \bm \cdot} + {r_k},1/(1 + {c_j^{(2)}})). \notag
\end{equation}
Similarly, %
we can expand the convolution along the other direction as
 ${m_{jkvl}\sim\mbox{Pois}(\sum_{f=1}^F d_{kvf}w_{jk(l-f+1)})}$, where ${w_{jk(l-f+1)}:=0}$ if ${l-f+1\notin \{1,2,\ldots,S_j\}}$, and obtain
 ${({\bm{d}}_{jkvl}\,|\, m_{jkvl})\sim \mbox{Multi}(m_{jkvl}; \xi_{jkvl1},\ldots,\xi_{jkvlF})}$, where ${\xi_{jkvlf}=d_{kvf}w_{jk(l-f+1)}/\sum_{f=1}^F d_{kvf}w_{jk(l-f+1)}}$ and ${\bm{d}}_{jkvl}^{{}} \in {{\mathbb{Z}}^{{F}}}$. %
Further applying the relationship between the Poisson and multinomial distributions, we have
\begin{equation}
\textstyle
((\bm d_{jk1\bm \cdot}',\ldots,\bm d_{jkV\bm \cdot}')' \,|\,m_{j k \bm \cdot \bm \cdot})\sim\mbox{Multi}(m_{j k \bm \cdot \bm \cdot}; \bm D_k(:)). \notag
\end{equation}
With the Dirichlet-multinomial conjugacy, we have
\begin{equation}
\textstyle
 ({{\bm{D}}_k(:)}\,|\,{-})\sim \mbox{Dir}((\bm d_{\bm \cdot k1\bm \cdot}',\ldots,\bm d_{\bm \cdot kV\bm \cdot}')'+\eta \bm 1_{|V|F}). \notag
\end{equation}

Exploiting the properties of the Poisson and multinomial distributions helps CPFA fully take advantages of the sparsity of the one-hot vectors, making its complexity comparable to a regular bag-of-words topic model that uses Gibbs sampling for inference.
Note as the multinomial related samplings inside each iteration are embarrassingly parallel, they are accelerated with GPUs in our experiments.

\begin{algorithm}
 \caption{Hybrid stochastic-gradient MCMC and autoencoding variational inference for CPGBN}
 {\small
 \begin{algorithmic}
 \STATE{Set mini-batch size $m$ and number of dictionaries ${K}$;}
 \STATE{Initialize encoder parameter ${\bm{\Omega}}$ and model parameter ${\{ {{\bm{D}}_{{k}}}\} _{1,{K}}}$;}
 \FOR{$iter = 1,2,\cdots$}
 \STATE{Randomly select a mini-batch of $m$ documents to form a subset ${\bm{X}} = {\{ {{\bm{X}}_j}\} _{1,m}}$;}
 \FOR{$j=1,2,\cdots$}
 \STATE{Draw random noise $ {\bm{\epsilon}} _j$ from uniform distribution;}
 \STATE{Calculate ${\nabla _{\bm{\Omega}} }L({\bm{\Omega}} ,{\bm{D}};{{\bm{x}}_j},{{\bm{\epsilon}} _j})$ according to (\ref{eq_lowerbound}) and update ${\bm{\Omega}}$;}
 \ENDFOR
 \STATE{Sample ${\{ {{\bm{w}}_j}\} _{1,m}}$ from (\ref{eq_w_post}) given ${\bm{\Omega}}$;}
 \STATE{Parallely process each positive point in ${\bm{X}}$ to obtain {$(\bm d_{\bm \cdot k1\bm \cdot}',\ldots,\bm d_{\bm \cdot k|V|\bm\cdot}')'$} according to (\ref{eq_w_aug})};
 \STATE{Update ${\{ {{\bm{D}}_{{k}}}\} _{1,{K}}}$ according to (\ref{eq_online})}
 \ENDFOR
 \end{algorithmic}
 \label{A1}}
\end{algorithm}

\subsection{{Hybrid} MCMC/Variational Inference}
\label{Hybrid}
While having closed-form update equations, the Gibbs sampler requires processing all documents in each iteration and hence has limited scalability.
Fortunately, there have been several related research on scalable inference for discrete LVMs (\citealp{ma2015a,patterson2013stochastic}).
Specifically, TLASGR-MCMC of \citealp{cong2017deep}, which uses an elegant simplex constraint and increases the sampling efficiency via the use of the Fisher information matrix (FIM), with adaptive step-sizes for the topics of different layers, can be naturally extended to our model.
The efficient TLASGR-MCMC update of ${\bm{D}}_{{k}}$ in CPFA can be described as
\begin{align}
&\resizebox{7.3cm}{!}{$\footnotesize \textstyle {\bm{D}}^{(new)}_k(:) = \Big\{{\bm{D}}_k(:) + \frac{{{\varepsilon_i}}}{{{{{M}}_k}}}[(\rho ({\bm{d}}_{\sim k1 \bm \cdot }',...,{\bm{d}}_{\sim k|V|\bm \cdot }')' + {\eta})$} \notag\\
&\resizebox{7.3cm}{!}{$\footnotesize -\textstyle (\rho {\bm{d}}_{\sim k\bm \cdot \bm \cdot } + {\eta}\left| V \right| F){\bm{D}}_k(:)] + N\big(0,\frac{{2{\varepsilon_i}}}{{{{{M}}_k}}}\mbox{diag}({\bm{D}}_k(:))\big)\Big\}_\angle$}, \label{eq_online}
\end{align}
where $i$ denotes the number of mini-batches processed so far; the symbol $\sim$ in the subscript denotes summing over the data in a mini-batch; and the definitions of $\rho$, $\varepsilon_i$, $\{\bm \cdot\}_\angle$, and ${{M}}_k$ are analogous to these %
in \citet{cong2017deep} and omitted here for brevity.

Similar to \citet{zhang2018whai}, combining TLASGR-MCMC and the convolutional inference network described in Section \ref{sec_ae}, we can construct a hybrid stochastic-gradient-MCMC/autoencoding variational inference for CPFA. More specifically, in mini-batch based each iteration, we draw a random sample of the CPFA global parameters ${\bm{D}} = {\{ {{\bm{D}}_{{k}}}\} _{1,{K}}}$ via TLASGR-MCMC; given the sampled global parameters, we optimize the parameters of the convolutional inference network, denoted as ${\bm{\Omega}}$, using the $-$ELBO in \eqref{eq:L}, which for CPFA is simplified as
\begin{equation}
 \begin{split}\textstyle
 {L_g} = -\sum_{j = 1}^J {{\mathbb{E}_{q({{\bm{w}}_j}\,|\,{{\bm{X}}_j})}}[\ln p({{\bm{X}}_j}\,|\,\{{\bm{D}}_k,{{\bm{w}}_{jk}}\}_{1,K})]} \\+ \textstyle\sum_{j = 1}^J {{\mathbb{E}_{q({{\bm{w}}_j}\,|\,{{\bm{x}}_j})}}[\ln \frac{{q({{\bm{w}}_j}\,|\,{{\bm{x}}_j})}}{{p({{\bm{w}}_j})}}]}.
 \label{eq_lowerbound}
 \end{split}
\end{equation}
We describe
the proposed hybrid stochastic-gradient-MCMC/autoencoding variational inference algorithm in Algorithm \ref{A1}, which is implemented in TensorFlow (\citealp{abadi2015tensorflow}), combined with pyCUDA (\citealp{klockner2012pycuda}) for more efficient computation.

\section{Related Work}
With the bag-of-words representation that ignores the word order information, a diverse set of deep topic models have been proposed to infer a multilayer data representation in an unsupervised manner.
A main mechanism of them is to connect adjacent layers by specific factorization, which usually boosts the performance (\citealp{gan2015scalable,zhou2016augmentable,zhang2018whai}).
However, limited by the bag-of-words representation, they usually perform poorly on sentiment analysis tasks, which heavily rely on the word order information (\citealp{xu2013convolutional,weston2014}).
In this paper, the proposed CPGBN could be seen as a novel convolutional extension, which not only clearly remedies the loss of word order, but also inherits various virtues of deep topic models.

Benefiting from the advance of word-embedding methods, CNN-based architectures have been leveraged as encoders for various natural language processing tasks (\citealp{kim2014, kalchbrenner2014a}).
They in general directly apply to the word embedding layer a single convolution layer, which, given a convolution filter window of size $n$, essentially acts as a detector of typical $n$-grams.
More complex deep neural networks taking CNNs as the their encoder and RNNs as decoder have also been studied for text generation (\citealp{zhang2016generating, semeniuta2017a}).
However, for unsupervised sentence modeling, language decoders other than RNNs are less well studied; it was not until recently that \citet{zhangdeconvolutional} have proposed a simple yet powerful, purely convolutional framework for unsupervisedly learning sentence representations, which is the first to force the encoded latent representation to capture the information from the entire sentence via a multi-layer CNN specification.
But there still exists a limitation in requiring an additional large corpus for training word embeddings, and it is also difficult to visualize and explain the semantic meanings learned by black-box deep networks.

For text categorization, the bi-grams (or a combination of bi-grams and unigrams) are confirmed to provide more discriminative power than unigrams (\citealp{tan2002the,glorot2011domain}).
Motivated by this observation, \citet{johnson2015effective} tackle document categorization tasks by directly applying shallow CNNs, with filter width three, on one-hot encoding document matrices, {outperforming both traditional $n$-grams and word-embedding based methods without the aid of additional training data.}
In addition, the shallow CNN serves as an important building block in many other supervised applications to help achieve sate-of-art results (\citealp{johnsonsemisupervised, johnson2017deep}).

\section{Experimental Results}
\begin{figure}
 \centering
 \includegraphics[width=6.5cm]{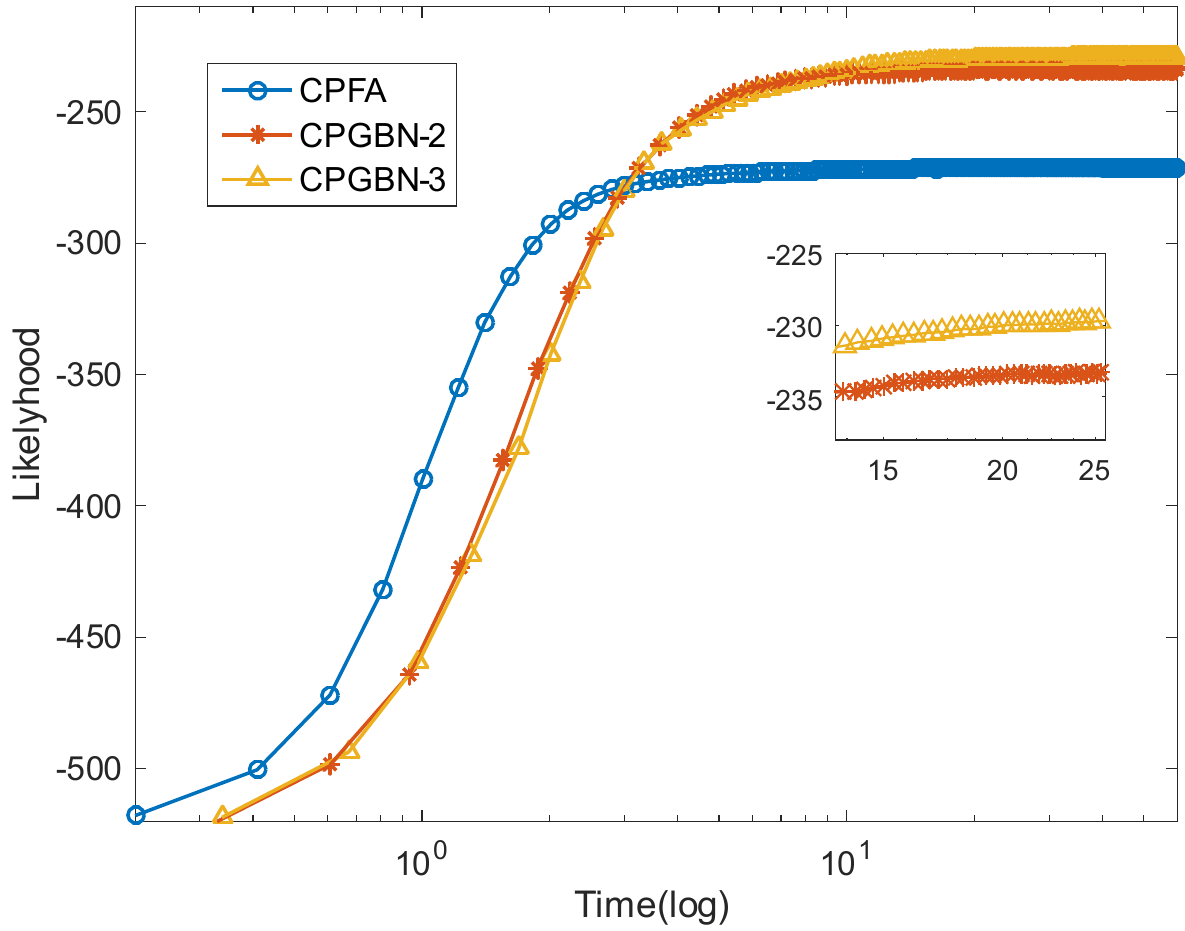} %
\vspace{-2mm}
 \caption{Point likelihood of CPGBNs on TREC as a function of time with various structural settings.}
\label{time_cmp} %
\end{figure}
\begin{figure}
 \centering
 \includegraphics[width=6.5cm]{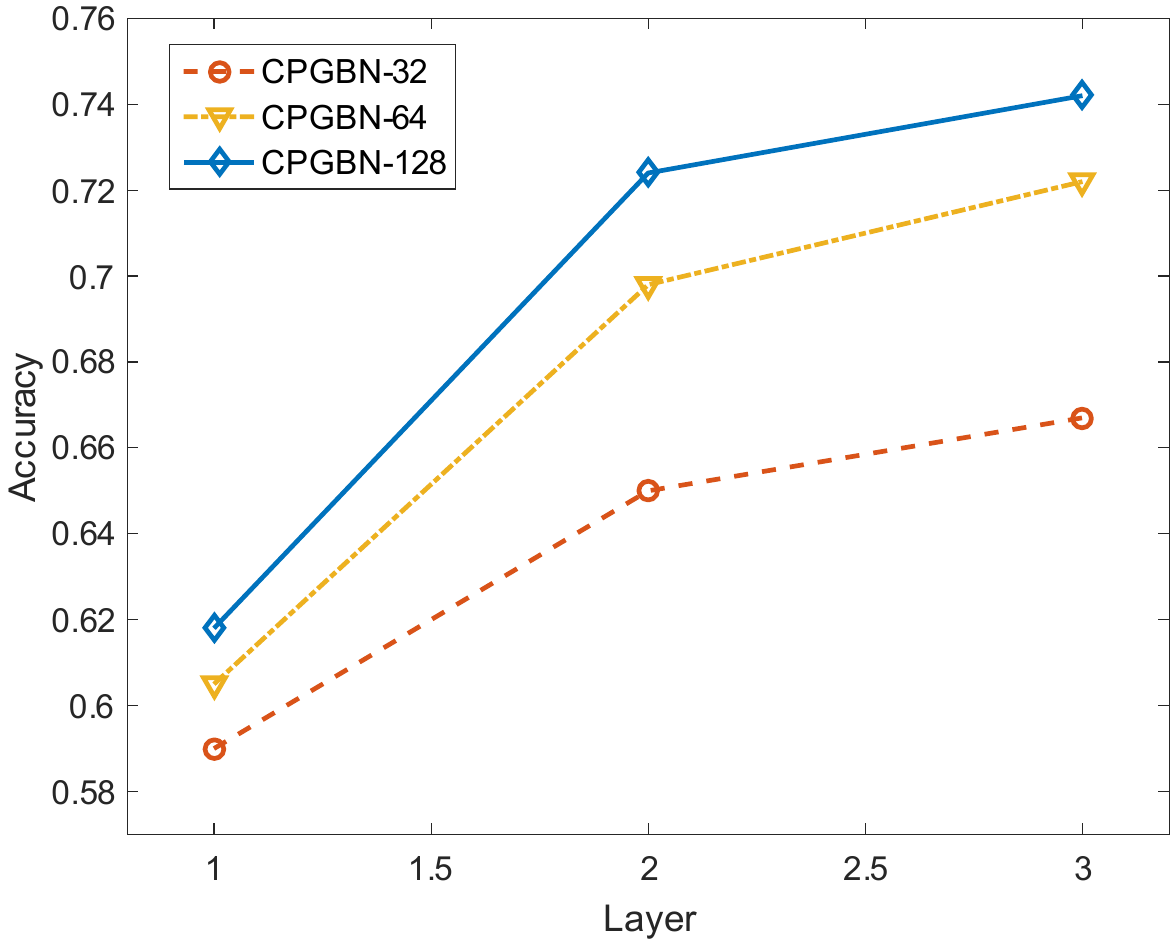} %
\vspace{-2mm}
 \caption{Classification accuracy $(\%)$ of the CPGBNs on TREC as a function of the depth with various structural settings.}
\label{ac_cmp} %
\end{figure}

\begin{table}[t]
 \centering
 \caption{Summary statistics for the datasets after tokenization ($C$: Number of target classes. $L$: Average sentence length. $N$: Dataset size. $V$: Vocabulary size. $V_{pre}$: Number of words present in the set of pre-trained word vectors. $Test$: Test set size, where CV means 10-fold cross validation).
 }
 \label{tab_data}
 \scalebox{0.95}{
 {
 \setlength{\tabcolsep}{2mm}{
 \begin{tabular}{|c||c|c|c|c|c|}
 \hline
 Data &MR &TREC &SUBJ &ELEC &IMDB \\
 \hline
 $C$ &2 &6 &2 &2 &2\\
 $L$ &20 &10 &23 &123 &266\\
 $N$ &10662 &5952 &10000 &50000 &50000\\
 $V$ &20277 &8678 &22636 &52248 &95212\\
 $V_{pre}$&20000 &8000 &20000 &30000 &30000\\
 $Test$ &CV &500 &CV &25000 &25000\\
 \hline
 \end{tabular}}}
 }
\end{table}

\begin{table*}[ht]
 \centering
 \caption{Comparison of classification accuracy on unsupervisedly extracted feature vectors and average training time (seconds per Gibbs sampling iteration across all documents) on three different datasets.
 }
 \label{tab_unsupervised}
 {
 \scalebox{0.85}{
 \setlength{\tabcolsep}{2.5mm}{
 \begin{tabular}{|c|c|ccc|ccc|}
 \hline
 \multirow{2}{*}{Model} &\multirow{2}{*}{Size} &\multicolumn{3}{|c|}{Accuracy} &\multicolumn{3}{|c|}{Time} \\
 \cline{3-8}
 & &MR &TREC &SUBJ &MR &TREC &SUBJ\\
 \hline
 \hline
 LDA &200 &54.4$\pm0.8$ &45.5$\pm1.9$ &68.2$\pm1.3$ &3.93 &0.92 &3.81\\
 \hline
 DocNADE &200 &54.2$\pm0.8$ &62.0$\pm0.6$ &72.9$\pm1.2$ &- &- &-\\
 \hline
 DPFA &200 &55.2$\pm1.2$ &51.4$\pm0.9$ &74.5$\pm1.9$ &6.61 &1.88 &6.53\\
 DPFA &200-100 &55.4$\pm0.9$ &52.0$\pm0.6$ &74.4$\pm1.5$ &6.74 &1.92 &6.62\\
 DPFA &200-100-50 &56.1$\pm0.9$ &62.0$\pm0.6$ &78.5$\pm1.4$ &6.92 &1.95 &6.80\\
 \hline
 PGBN &200 &56.3$\pm0.6$ &66.7$\pm1.8$ &76.2$\pm0.9$ &3.97 &1.01 &3.56\\
 PGBN &200-100 &56.7$\pm0.8$ &67.3$\pm1.7$ &77.3$\pm1.3$ &5.09 &1.72 &4.39\\
 PGBN &200-100-50 &57.0$\pm0.5$ &67.9$\pm1.5$ &78.3$\pm1.2$ &5.67 &1.87 &4.91\\
 \hline
 WHAI &200 &55.6$\pm0.8$ &60.4$\pm1.9$ &75.4$\pm1.5$ &- &- &-\\
 WHAI &200-100 &56.2$\pm1.0$ &63.5$\pm1.8$ &76.0$\pm1.4$ &- &- &-\\
 WHAI &200-100-50 &56.4$\pm0.6$ &65.6$\pm1.7$ &76.5$\pm1.1$ &- &- &-\\
 \hline
 CPGBN &200 &61.5$\pm0.8$ &68.4$\pm0.8$ &77.4$\pm0.8$ &3.58 &0.98 &3.53\\
 CPGBN &200-100 &62.4$\pm0.7$ &73.4$\pm0.8$ &81.2$\pm0.8$ &8.19 &1.99 &6.56\\
 CPGBN &200-100-50 &\textbf{63.6}$\pm0.8$ &\textbf{74.4}$\pm0.6$ &\textbf{81.5}$\pm0.6$ &10.44 &2.59 &7.87\\
 \hline
 \end{tabular}
 }
 }}
 \vspace{-3mm}
\end{table*}
\begin{table*}
 \centering
 \caption{Example phrases learned from TREC by CPGBN.}
 \label{tab_phase}
 {
 \scalebox{0.84}{
 \setlength{\tabcolsep}{3mm}{
 \begin{tabular}{|c|c|c|c|c|}
 \hline
 \multirow{2}{*}{Kernel Index } &\multicolumn{3}{|c|}{Visualized Topic} &\multirow{2}{*}{Visualized Phrase } \\
 \cline{2-4}
 &1st Column & 2nd Column &3rd Column & \\
 \hline
 \multirow{4}{*}{192th Kernel} &\textbf{how} & do & you
 &\multirow{4}{3.5cm}{\textbf{how} do you,\\ \textbf{how} many years,\\ \textbf{how} much degrees}
 \\
 & cocktail & many & years &\\
 & stadium & much & miles &\\
 & run & long & degrees &\\
 \hline
 \multirow{4}{*}{80th Kernel} &microsoft &e-mail &address
 &\multirow{4}{3.5cm}{microsoft e-mail address, microsoft email address, virtual ip address}
 \\
 & virtual & email & addresses &\\
 & answers.com & ip & floods &\\
 & softball & brothers & score &\\
 \hline
 \multirow{4}{*}{177th Kernel} &\textbf{who} &created &maria
 &\multirow{4}{3.5cm}{\textbf{who} created snoopy,\\ \textbf{who} fired caesar,\\ \textbf{who} wrote angela}
 \\
 & willy & wrote & angela &\\
 & bar & fired & snoopy &\\
 & hydrogen & are & caesar &\\
 \hline
 \multirow{4}{*}{47th Kernel} &dist &\textbf{how} &far
 &\multirow{4}{3.5cm}{dist \textbf{how} far,\\ dist \textbf{how} high , \\dist \textbf{how} tall}
 \\
 & all-time & stock & high &\\
 & wheel & 1976 & tall &\\
 & saltpepter & westview & exchange &\\
 \hline
 \end{tabular}}}}
 \vspace{-2mm}
\end{table*}

\subsection{Datasets and Preprocessing}
We test the proposed CPGBN and its supervised extension (sCPGBN) on various benchmarks, including:

$\bullet$ $\bm{{\rm{MR}}}$: Movie reviews with one sentence per review, where the task is to classify a review as being positive or negative
(\citealp{pang2005seeing}).%

$\bullet$ $\bm{{\rm{TREC}}}$: TREC question dataset, where the task is to classify a question into one of six question types (whether the question is about abbreviation, entity, description, human, location, or numeric)
(\citealp{li2002learning}).%

$\bullet$ $\bm{{\rm{SUBJ}}}$: Subjectivity dataset, where the task is to classify a sentence as being subjective or objective (\citealp{pang2004ab}).

$\bullet$ $\bm{{\rm{ELEC}}}$: ELEC dataset (\citealp{mcauley2013hidden}) consists of electronic product reviews, which is part of a large Amazon review dataset.%

$\bullet$ $\bm{{\rm{IMDB}}}$: IMDB dataset (\citealp{maas2011learning}) is a benchmark dataset for sentiment analysis, where the task is to determine whether a movie review is positive or negative.%

We follow the steps listed in \citet{johnson2015effective} to tokenize the text, where emojis such as ``:-)'' are treated as tokens and all the characters are converted to lower case.
We then select the top $V_{pre}$ most frequent words to construct the vocabulary, without dropping stopwords; we map the words not included in the vocabulary to a same special token to keep all sentences structurally intact.
The summary statistics of all benchmark datasets are listed in Table~\ref{tab_data}.

\subsection{Inference Efficiency}

In this section we show the results of the proposed CPGBN on TREC.
First, to demonstrate the advantages of {increasing the depth of the network,} we construct three networks of different depths: with ${(K=32)}$ for $T=1$, $(K_1=32, K_2=16)$ for $T=2$, and $(K_1=32,K_2=16, K_3=8)$ for $T=3$. Under the same configuration of filter width $F = 3$ and the same hyperparameter setting, where ${e_0} = {f_0} = 0.1$ and $\eta^{(t)} = 0.05$, the networks are trained with the proposed Gibbs sampler.
The trace plots of model likelihoods are shown in Fig. \ref{time_cmp}.
It is worth noting that increasing the network depth in general improves the quality of data fitting, but as the complexity of the model increases, the model tends to converge more slowly in time.

Considering that the data fitting and generation ability is not necessarily strongly correlated the performance on specific tasks, we evaluate the proposed models on document classification.
Using the same experimental settings as mentioned above, we investigate how the classification accuracy is impacted by the network structure.
On each network, we apply the Gibbs sampler to collect 200 MCMC samples after 500 burn-ins to estimate the posterior mean of the feature usage weight vector ${\bm{w}}_j$, for every document in both the training and testing sets.
A linear support vector machine (SVM) (\citealp{cortes1995support-vector}) is taken as the classifier
{on the first hidden layer, denoted as ${\bm{{\theta}} _j^{(1)}}$ in \eqref{eq_PGCN}, to make a fair comparison}, where each result listed in Table \ref{tab_unsupervised} is the average accuracy of five independent runs.
Fig. \ref{ac_cmp} shows a clear trend of improvement in classification accuracy, by increasing the network depth given a limited first-layer width, or by increasing the hidden-layer width given a fixed depth.

\subsection{Unsupervised Models}
In our second set of experiments, we evaluate the performance of different unsupervised algorithms on MR, TREC, and SUBJ datasets by comparing the discriminative ability of their unsupervisely extracted latent features.
We consider LDA (\citealp{blei2003latent}) and its deep extensions, including DPFA (\citealp{gan2015scalable}) and PGBN \cite{zhou2016augmentable}, which are trained with batch Gibbs sampling.
We also consider WHAI (\citealp{zhang2018whai}) and DocNADE (\citealp{lauly2017document}) that are trained with stochastic gradient descent.

To make a fair comparison, we let CPGBNs to have the same hidden layer widths as the other methods, and set the filter width as 3 for the convolutional layer.
Listed in Table~\ref{tab_unsupervised} are the results of various algorithms, where the means and error bars are obtained from five independent runs, using the code provided by the original authors.
For all batch learning algorithms, we also report in Table \ref{tab_unsupervised} their average run time for an epoch ($i.e.,$ processing all training documents once).
Clearly, given the same generative network structure, CPGBN performs the best in terms of classification accuracy, which can be attributed to its ability to utilize the word order information.
The performance of CPGBN has a clear trend of improvement as the generative network becomes deeper, which is also observed on other deep generative models including DPFA, PGBN, and WHAI.
In terms of running time, the shallow LDA could be the most efficient model compared to these more sophisticated ones, while CPGBN of a single layer achieves a comparable effectiveness thanks to its efficient use of GPU for parallelizing its computation inside each iteration.
Note all running times are reported based {on} a Nvidia GTX 1080Ti GPU.

In addition to quantitative evaluations, we have also visually inspected the inferred convolutional kernels of CPGBN, which is distinct from many existing convolutional models that build nonlinearity via ``black-box'' neural networks.
As shown in Table \ref{tab_phase}, we list several convolutional kernel elements of filter width 3 learned from TREC, using a single-hidden-layer CPGBN of size 200.
We exhibit the top 4 most probable words in each column of the corresponding kernel element.
It's particularly interesting to note that the words in different columns can be combined into a variety of interpretable phrases with similar semantics.
CPGBN explicitly take the word order information into consideration to extract phrases, which are {then} combined into a hierarchy of phrase-level topics, helping clearly improve the quality of unsupervisedly extracted features.
 Take the 177th convolutional kernel for example, the top word of its 1st topic is “who,” its 2nd topic is a verb topic: “created, wrote, fired, are,'' while its 3rd topic is a noun topic: ``maria/angela/snoopy/caesar.'' These word-level topics can be combined to construct phrases such as ``who, created/ wrote/ fired/are, maria/angela/snoopy/caesar,'' resulting in a phrase-level topic about ``human,'' one of the six types of questions in TREC. Note these shallow phrase-level topics will become more general in a deeper layer of CPGBN. %
 We provide two example phrase-level topic hierarchies in the Appendix to enhance interpretability.

\subsection{Supervised Models}

Table \ref{tab_supervised} lists the comparison of various supervised algorithms on three common benchmarks, including SUBJ, ELEC, and IMDB.
The results listed there are either quoted from published papers, or reproduced with the code provided by the original authors.
We consider bag-of-words representation based supervised topic models, including sAVITM (\citealp{srivastava2017autoencoding}), MedLDA (\citealp{zhu2014gibbs}), and sWHAI (\citealp{zhang2018whai}).
{We also consider three types of bag-of-$n$-gram models (\citealp{johnson2015effective}), where $n \in \{ 1,2,3\}$}, and word embedding based methods, indicated with suffix ``-wv,'' including SVM-wv (\citealp{zhang2017a}) and RNN-wv and LSTM-wv (\citealp{johnson2016supervised}).
In addition, we consider several related CNN based methods, including three different variants of Text CNN (\citealp{kim2014})---CNN-rand, CNN-static, and CNN-non-static---%
and CNN-one-hot (\citealp{johnson2015effective}) that is based on one-hot encoding.

We construct three different sCPGBNs with $T \in \{1,2,3\}$, as described in Section \ref{sec_ae}.
As shown in Table \ref{tab_supervised}, the word embedding based methods generally outperform the methods based on bag-of-words, which is not surprising as the latter completely ignore word order.
Among all bag-of-words representation based methods, sWHAI performs the best {and even} achieves comparable performance to some word-embedding based methods, which illustrates the benefits of having multi-stochastic-layer latent representations.
As for $n$-grams based models, although they achieve comparable performance to word-embedding based methods, we find via experiments that both their performance and computation are sensitive to the vocabulary size. %
Among the CNN related algorithms, CNN-one-hot tends to have a better performance on classifying longer texts than Text CNN does, which agrees with the observations of \citet{zhang2017a}; possible explanation for this phenomenon is that CNN-one-hot is prone to overfitting on short documents.
Moving beyond CNN-one-hot, sCPGBN could help capture the underlying high-order statistics to alleviate overfitting, as commonly observed in deep generative models (DGMs) (\citealp{li2015max-margin}), and improves its performance by increasing its number of stochastic hidden layers.

\begin{table}
 \centering
 \caption{Comparison of classification accuracy on supervised feature extraction tasks on three different datasets.}
 \label{tab_supervised}
 \scalebox{0.86}
 {\small
 \setlength{\tabcolsep}{0.75mm}{
 \begin{tabular}{|c|ccc|}
 \hline
 Model &SUBJ &ELEC &IMDB\\
 \hline
 sAVITM \citep{srivastava2017autoencoding} &85.7 &83.7 &84.9\\
 MedLDA \citep{zhu2014gibbs} &86.5 &84.6 &85.7\\
 sWHAI-layer1 \citep{zhang2018whai} &90.6 &86.8 &87.2 \\
 sWHAI-layer2 \citep{zhang2018whai} &91.7 &87.5 &88.0 \\
 sWHAI-layer3 \citep{zhang2018whai} &92.0 &87.8 &88.2\\
 \hline
 {SVM-unigrams} \citep{tan2002the} &88.5 &86.3 &87.7\\
 {SVM-bigrams} \citep{tan2002the} &89.4 &87.2 &88.2\\
 {SVM-trigrams} \citep{tan2002the} &89.7 &87.4 &88.5\\
 \hline
 SVM-wv \citep{zhang2017a} &90.1 &85.9 &86.5\\
 RNN-wv \citep{johnson2016supervised} &88.9 &87.5 &88.3\\
 LSTM-wv \citep{johnson2016supervised} &89.8 &88.3 &89.0\\
 \hline
 CNN-rand \citep{kim2014} &89.6 &86.8 &86.3\\
 CNN-static \citep{kim2014} &93.0 &87.8 &88.9\\
 CNN-non-static \citep{kim2014} &93.4 &88.6 &89.5\\
 CNN-one-hot \citep{johnson2015effective} &91.1 &91.3 &91.6\\
 \hline
 sCPGBN-layer1 &93.4$\pm0.1$ &91.6$\pm0.3$ &91.8$\pm0.3$\\
 sCPGBN-layer2 &93.7$\pm0.1$ &92.0$\pm0.2$ &92.4$\pm0.2$\\
 sCPGBN-layer3 &\textbf{93.8}$\pm0.1$ &\textbf{92.2}$\pm0.2$ &\textbf{92.6}$\pm0.2$\\
 \hline
 \end{tabular}
 }}
 \vspace{-6mm}
\end{table}

\section{Conclusion}
We propose convolutional Poisson factor analysis (CPFA), a hierarchical Bayesian model that represents each word in a document as a one-hot vector, and captures the word order information by performing convolution on sequentially ordered one-hot word vectors.
By developing a principled document-level stochastic pooling layer, we further couple CPFA with a multi-stochastic-layer deep topic model to construct convolutional Poisson gamma belief network (CPGBN).
We develop a Gibbs sampler to jointly train all the layers of CPGBN.
For more scalable training and fast testing, we further introduce a mini-batch based stochastic inference algorithm that combines both stochastic-gradient MCMC and a Weibull distribution based convolutional variational auto-encoder. In addition, we provide a supervised extension of CPGBN.
Example results on both unsupervised and supervised feature extraction tasks show CPGBN combines the virtues of both convolutional operations and deep topic models, providing not only state-of-the-art classification performance, but also highly interpretable phrase-level deep latent representations.

\section*{Acknowledgements}
B. Chen acknowledges the support of the Program for Young Thousand Talent by Chinese Central Government, the 111 Project (No. B18039), NSFC (61771361), NSFC for Distinguished Young Scholars (61525105), and the Innovation Fund of Xidian University. M. Zhou acknowledges the support of Award IIS-1812699 from the U.S. National Science Foundation and the McCombs Research Excellence Grant.

\bibliography{example_paper,References052016,reference}
\bibliographystyle{icml2019}
%\end{document}
$ $
\newpage
$ $
\normalsize

\newpage
\appendix

\section{Inference for CPGBN}
Here we describe the derivation in detail for convolutional Poisson gamma belief network (CPGBN) with $T$ hidden layers, expressed as
\begin{equation}
 \small
 \begin{array}{l}
 {\bm{\theta}} _j^{(T)}\sim \mbox{Gam}(\bm{r},1/c_j^{(T + 1)}),\\
 ...,\\
 {\bm{\theta}} _j^{(t)}\sim \mbox{Gam}({{\bm{\Phi}} ^{(t + 1)}}{\bm{\theta}} _j^{(t + 1)},1/c_j^{(t + 1)}),\\
 {...,}\\
 {\bm{\theta}} _j^{(1)}\sim\mbox{Gam}({{\bm{\Phi}} ^{(2)}}{\bm{\theta}} _j^{(2)},1/c_j^{(2)}),\vspace{1.1mm}\\
 {\bm{w}}_{j{k}} = {\bm{\pi}_{jk}} \theta _{j{k}}^{(1)},~{\bm{\pi}_{jk}} \sim \mbox{Dir}\big( {\bm{\Phi}} _{{k:}}^{({\rm{2}})}{\bm{\theta}} _j^{({\rm{2}})} /S_j \bm{1}_{S_j}\big),\vspace{1.1mm}\\
 {\bm{M}}_j\sim \mbox{Pois}\big(\sum\nolimits_{{k} = 1}^{{K^{(1)}}} {{\bm{D}}_{{k}}*{\bm{w}}_{j{k}}} \big).
 \end{array}
 \normalsize
 \label{appendix_model}
\end{equation}
Note using the relationship between the gamma and Dirichlet distributions ($e.g.$, Lemma IV.3 of \citet{zhou2012negative}), the elements of $\bm{w}_{j{k}}$ in the first hidden layer can be equivalently generated as \begin{equation}
 \begin{array}{l}
 w_{j{k}s}\sim \mbox{Gam}\big( {\bm{\Phi}} _{{k:}}^{(2)}{\bm{\theta}} _j^{(2)}/S_j,1/c_j^{(2)}\big),~s=1,\ldots,S_j.
 \end{array}\!\!\!
 \label{appendix_pool}
\end{equation}
Note the random variable $\theta _{j{k^{}}}^{(1)}$, which pools the random weights of all words in document $j$, follows
\begin{equation}
\!\!\!
\begin{array}{l}
\theta _{j{k^{}}}^{(1)}{=}\sum\nolimits_{s^{} = 1}^{S_j^{}} {w_{j{k^{}}s^{}}^{}}
\sim\mbox{Gam}({\bm{\Phi}} _{{k^{}}:}^{(2)}{\bm{\theta}} _j^{(2)},1/c_j^{(2)}).\\
\end{array}
\end{equation}
As described in Section 3.1, we have %
\begin{equation}
\begin{array}{l}
\bm m_{j k \bm \cdot \bm \cdot} \sim \mbox{Pois}({{\bm{w}}_{jk}}),\\
\end{array}
\label{appendix_like}
\end{equation}
\begin{equation}
\small
\begin{array}{l}
((\bm d_{jk1\bm \cdot}',\ldots,\bm d_{jkV\bm \cdot}')' \,|\,m_{j k \bm \cdot \bm \cdot})\sim\mbox{Multi}(m_{j k \bm \cdot \bm \cdot}; \bm D_k(:))
\end{array}
\normalsize
\end{equation}
leading to the following conditional posteriors:
 \begin{equation}\textstyle
 ({{\bm{w}}_{jk}}\,|\,{-})\sim \mbox{Gam}(\bm m_{j k \bm \cdot \bm \cdot} + {r_k},1/(1 + {c_j^{(2)}})). \notag
\end{equation}
\begin{equation}
\begin{array}{l}
 ({{\bm{D}}_k(:)}\,|\,{-})\sim \mbox{Dir}((\bm d_{\bm \cdot k1\bm \cdot}',\ldots,\bm d_{\bm \cdot kV\bm \cdot}')'+\eta \bm 1_{|V|F}).
\end{array}
\normalsize
\end{equation}

Since ${\bm{w}}_{j{k^{}}}^{}{\rm{ = }}{\bm{\pi}} _{j{k^{}}}^{}\theta _{j{k}}^{(1)}$, %
 from \eqref{appendix_like} we have %
\begin{equation}
\begin{split}
{m_{j{k^{}}\bm \cdot \bm \cdot s^{}}^{}\sim \mbox{Pois}(\pi _{j{k^{}}s^{}}^{}\theta _{j{k^{}}}^{(1)})} ,
\end{split}
\end{equation}
\begin{equation}
\begin{split}
{m_{j{k^{}}\bm \cdot \bm \cdot \bm \cdot^{}}^{}} \sim \mbox{Pois}(\theta _{j{k^{}}}^{\left( 1 \right)})
\end{split}.
\end{equation}
Since $\sum_{s=1}^{S_j}\pi_{jks} =1$ by construction, we have
\begin{equation}
\begin{split}
\!\!\!\!
( {{\bm{m}}_{j{k^{}}\bm \cdot \bm \cdot}^{}} \,| \,{m_{j{k^{}}\bm \cdot \bm \cdot \bm \cdot ^{}}^{}} )\sim \mbox{Multi}({m_{j{k^{}}\bm \cdot \bm \cdot \bm \cdot ^{}}^{}};{\bm{\pi}} _{j{k^{}}}^{}),
\end{split}
\end{equation}
and hence the following conditional posteriors:
\begin{equation}
\begin{split}
\!\!\!\!\!
(\theta _{j{k^{}}}^{\left( 1 \right)}\,|\,-)\sim \mbox{Gam}&( {m_{j{k^{}}\bm \cdot \bm \cdot \bm \cdot^{}}^{}} + {\bm{\Phi}} _{{k^{}:}}^{\left( 2 \right)}{\bm{\theta}} _j^{\left( 2 \right)},1/(1 + c_j^{(2)})),
\end{split}
\end{equation}
\begin{equation}
\begin{split}
\!\!\!\!\!\!\!\!\!\!\!\!\!\!\!\!\!\!\!\!\!\!\!\!\!
({\bm{\pi}} _{j{k^{}}}^{} |-) \sim
\mbox{Dir}&( {{\bm{m}}_{j{k^{}}\bm \cdot \bm \cdot}^{}} / {m_{j{k^{}}\bm \cdot \bm \cdot \bm \cdot^{}}^{}} +
{{{\bm{\Phi}} _{{k^{}:}}^{\left( 2 \right)}{\bm{\theta}} _j^{\left( 2 \right)}}/{S_j^{}}}),
\end{split}
\end{equation}
\begin{equation}
\small
\begin{split}
\!\!\!
\textstyle(c_j^{(2)}| - )\sim\mbox{Gam}(&\textstyle\sum_{{k^{}} = 1}^{{K^{(1)}}} {{\bm{\Phi}} _{{k^{}}:}^{(2)}{\bm{\theta}} _j^{(2)}}{+}{a_0},1/(\sum_{{k^{}} = 1}^{{K^{(1)}}} {\theta _{j{k^{}}}^{(1)}}{+}{b_0})).
\end{split}
\normalsize
\end{equation}
The derivation for the parameters of layer $t \in \{ 2,...,T\}$ is the same as that of gamma belief network (GBN) \cite{zhou2016augmentable}, omitted here for brevity.

\section{Sensitivity to Filter Width}

To investigate the effect of the filter width of the convolutional kernel, we have evaluated the performance of CPFA ($i.e.$, CPGBN with a single hidden layer) on the SUBJ dataset with a variety of filter widths (unsupervised feature extraction + linear SVM for classification). We use the same CPFA code  but vary its setting of the filter width.  Averaging over five independent runs, the accuracy for filter wdith 1, 2, 3, 4, 5, 6, and 7 are $74.9 \pm 0.9$, $77.3 \pm 0.4$, $77.5 \pm 0.5$, $77.8 \pm 0.4$, $77.6 \pm 0.5$, $78.0 \pm 0.4$, and $77.5 \pm 0.4$, respectively. Note when the filter width reduces to 1, CPFA reduces to PFA ($i.e.$, no convolution). These results suggest the performance of CPFA has low sensitivity to the filter width. While setting the filter width as three may not be the optimal choice, it is a common practice for existing text CNNs \citep{kim2014, johnson2015effective}. 

\section{Hierarchical Visualization}
\begin{figure*}
 \centering
 \includegraphics[height=7.2cm,width=16cm]{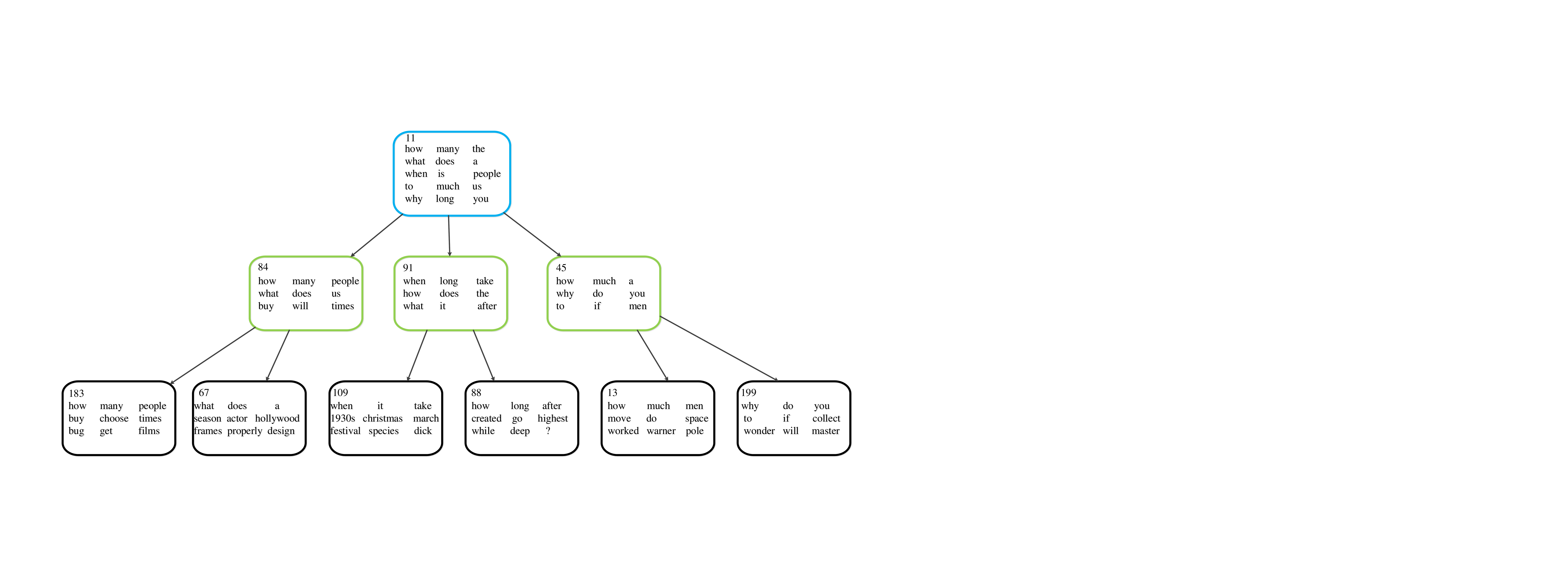} %
 \caption{The $[6,3,1]$ phrase-level tree that includes all the lower-layer nodes (directly or indirectly) linked %
 to the $11th$ node of the top layer, taken from the full $[200,100,50]$ network inferred by CPGBN on TREC dataset.}
\label{hir_vis} %
\end{figure*}

\begin{figure*}
 \centering
 \includegraphics[height=7.2cm,width=16cm]{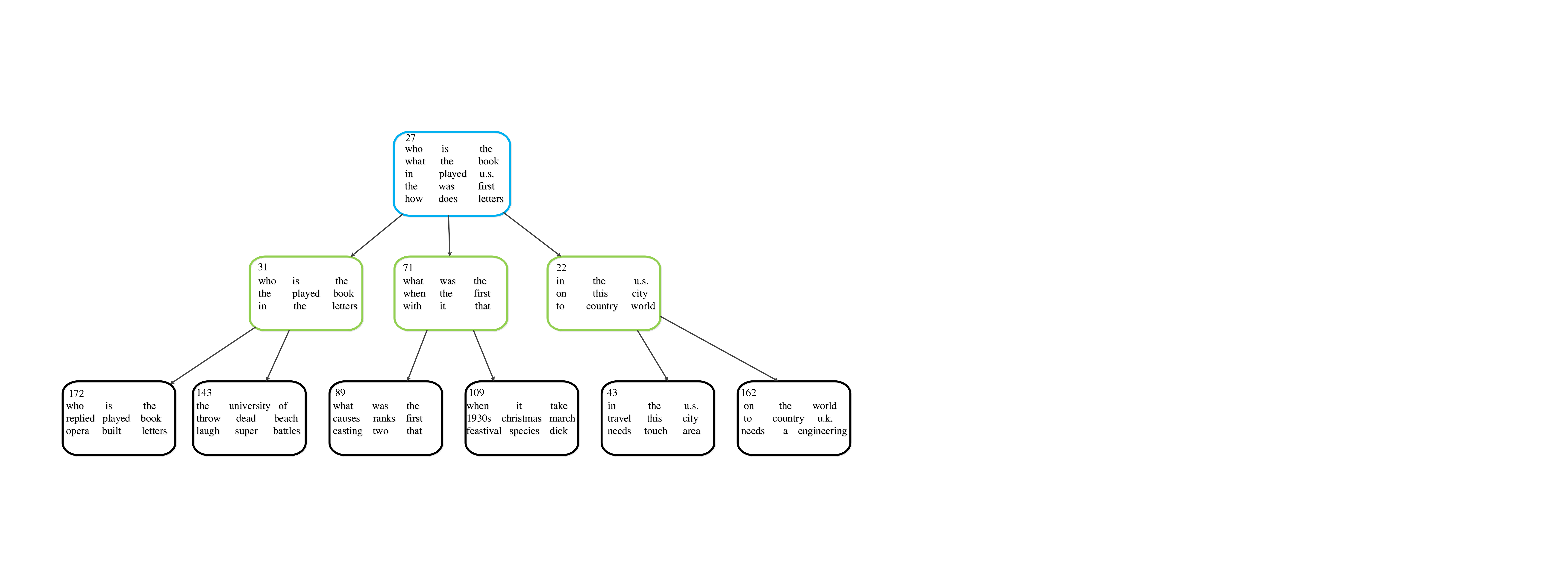} %
 \caption{The $[6,3,1]$ phrase-level tree that include all the lower-layer nodes (directly or indirectly) linked %
 to the $27th$ node of the top layer, taken from the full $[200,100,50]$ network inferred by CPGBN on TREC dataset.}
\label{hir_vis} %
\end{figure*}

Distinct from word-level topics learned by traditional topic models \cite{LSA,LSI,NMF,blei2003latent,hinton2009replicated,BNBP_PFA_AISTATS2012}, we propose novel phrase-level topics preserving word order as shown in Table. 3, where each phrase-level topic is often combined with several frequently co-occurred short phrases. To explore the connections between phrase-level topics of different layers learned by CPGBN, we follow \citet{zhou2016augmentable} to construct trees to understand the general and specific aspects of the corpus. More specifically, we construct trees learned from TREC dataset, with the network structure set as $[K^{(1)},K^{(2)},K^{(3)}] = [200,100,50]$. We pick a node at the top layer as the root of a tree and grow the tree downward by drawing a line from node $k$ at layer $t$ %
to the top $M$ relevant nodes $k'$ at layer $t-1$.

As shown in Fig. \ref{hir_vis}, we select the top 3 relevant nodes at the second layer linked to the selected root node, and the top 2 relevant nodes at the third layer linked to the selected nodes at the second layer. Considering the TREC corpus only consists of questions (questions about abbreviation, entity, description, human, location, or numeric), most of the topics learned by CPGBN are focused on short phrases on asking specific questions, as shown in Table. 3. Following the branches of the tree in Fig. \ref{hir_vis}, the root node covers very general question types on ``how many, how long, what, when, why," and it is clear that the topics become more and more specific when moving along the tree from the top to bottom, where the shallow topics of the first layer tend to focus on a single question type, $e.g.$, the $183th$ bottom-layer node queries ``how many" and the $88th$ one queries ``how long."
\end{document}